\documentclass[runningheads]{llncs}
\usepackage[T1]{fontenc}
\usepackage{graphicx}
\usepackage{booktabs}
\usepackage[misc]{ifsym}

\usepackage{multirow}
\usepackage{xcolor}
\usepackage{hyperref}
\usepackage{amsmath}
\usepackage{amssymb}

\begin{document}

\author{Adeel Zafar \and
S\l{}awomir Nowaczyk 
}
\authorrunning{A. Zafar et al.}
\institute{Center for Applied Intelligent Systems Research, Halmstad University, Sweden\\
\email{\{adeel.zafar, slawomir.nowaczyk\}@hh.se}}

\title{Mind the Gap: Robustness Risks in PII Detection Systems}

\maketitle

\begin{abstract}
Personally Identifiable Information (PII) detection is a foundational component of data protection infrastructure where missed entities constitute direct privacy and security risks. Although modern PII systems report strong performance on standard benchmarks, we show that these evaluations mask substantial robustness failures under realistic distribution shifts encountered in deployment. Rather than comparing state-of-the-art accuracy, we study how different PII detection paradigms fail under noisy, unstructured, and informal inputs. We construct a stress test benchmark spanning seven categories of natural distribution shift and evaluate representative systems from three widely deployed architectural families: encoder-based NER (SpaCy), rule-based hybrid detection (Presidio), and generative LLM extraction (Qwen2.5-3B). 

All three exhibit significant degradation on out-of-distribution inputs, but with distinct and complementary failure modes. Encoder models primarily fail on unseen surface forms and boundary detection, rule-based systems fail on non-standard formats, and LLMs exhibit entity-type confusion and generation instability. These results show that aggregate benchmark scores obscure deployment-critical weaknesses and that no single architecture is uniformly reliable across PII categories. Motivated by these findings, we propose a hybrid detection pipeline with a QA-driven feedback loop for iterative risk mitigation, and release our benchmark to support OOD-aware evaluation of PII systems.

\end{abstract}

%==============================================================================
\section{Introduction}
\label{sec:introduction}
%==============================================================================

PII detection is a safety critical task in modern NLP pipelines and a foundational component of data protection infrastructure. Organizations rely on these systems to identify and redact sensitive entities before data is stored, shared, or processed. A missed entity is not merely an accuracy shortfall but a data protection failure with consequences under regulations such as GDPR~\cite{voigt2017gdpr}. As organizations increasingly automate PII redaction in pipelines handling customer data, medical records, and financial documents, the reliability of these systems becomes a direct security concern. Current PII detection systems, whether encoder based~\cite{devlin2019bert}, rule-based Presidio\footnote{Presidio: Data Protection and De-identification. Available at: \url{https://microsoft.github.io/presidio/}. Accessed: July 2026.}, or LLM based~\cite{brown2020language}, report F1 scores exceeding 90\% on standard benchmarks~\cite{tjong2003conll,weischedel2013ontonotes}. However, this confidence is misleading because these benchmarks consist of well edited, formally structured text drawn from news articles and encyclopedic sources, while real world inputs are messy, unstructured, and highly variable across domains.

The distribution shift problem is compounded when PII detection systems are deployed across different verticals. A model trained on general purpose corpora encounters fundamentally different language patterns in healthcare records, financial documents, customer support transcripts, or social media. The entity types may be the same but the surrounding context, formatting conventions, and vocabulary differ substantially. These cross domain shifts interact with the textual messiness we study in this paper, creating compounding challenges that current benchmarks do not capture.

Our goal is not to identify a single state-of-the-art PII detector, but to study how fundamentally different detection paradigms fail under realistic distribution shifts. We therefore evaluate representative systems from three widely deployed architectural families: encoder-based NER, rule-based hybrid detection, and generative LLM extraction. By comparing qualitatively different detection mechanisms rather than optimizing leaderboard performance, we aim to characterize complementary robustness failures that aggregate benchmark scores obscure.

In our evaluation, we find that all three architectural families exhibit significant and systematic performance degradation on unstructured inputs. Flexible format entities such as \textsc{Location} and \textsc{Address} suffer the most, with \textsc{Location} recall dropping from 0.725 to 0.457 across all models, while rigid format entities like SSN and IP addresses remain largely unaffected. Crucially, each architecture fails in different ways on the same input. Consider the following example:

\begin{quote}
\textit{``name William Stevenson address 588 Erickson Hills Suite 055 South Brandytown, PA ssn 486-34-9478 dob 08/21/1981''}
\end{quote}

\noindent The encoder model detects the person name but misses the location. The rule based hybrid detects the person, SSN, and date of birth but also misses the location. The LLM detects most entities but absorbs the city into a single ADDRESS span rather than extracting it separately. Each architecture fails for a different internal reason yet PII remains exposed in the output.

A natural question is why evaluate encoder and rule based models when LLMs increasingly dominate NLP tasks. Our results provide a clear answer: no single architecture is universally superior for PII detection. The LLM achieves the best overall F1 but shows lower PERSON recall than SpaCy (0.777 vs 0.951), worse CREDIT\_CARD detection than Presidio (0.543 vs 0.700), produces unparseable outputs on 2.5\% of OOD inputs, and is orders of magnitude slower and more expensive to run at scale. Meanwhile, Presidio's regex patterns achieve perfect SSN and IP detection that the LLM cannot match, and SpaCy provides reliable, fast person name detection. Understanding these tradeoffs is essential for building production PII systems that balance accuracy, latency, cost, and reliability.

In this paper, we make three contributions: 
(1) we characterize architecture-specific failure modes in PII detection under realistic distribution shifts, showing that encoder, rule-based, and LLM systems fail in distinct and complementary ways; 
(2) we construct a stress test benchmark\footnote{ Source Code: \url{https://github.com/Adeelzafar/Mind-the-Gap-PII}} spanning seven categories of natural distribution shift; and 
(3) we argue for OOD-aware evaluation and hybrid detection pipelines that combine complementary architectural strengths.

%==============================================================================
\section{Related Work}
\label{sec:related}
%==============================================================================

\subsection{Named Entity Recognition and PII Detection}

Named Entity Recognition (NER) has evolved through several architectural paradigms. Early statistical approaches using CRFs and HMMs~\cite{rustad2024systematic} were followed by neural architectures combining BiLSTMs with CRF layers. The introduction of pretrained transformer models marked a significant shift, with BERT~\cite{devlin2019bert} and its variants such as DeBERTa~\cite{he2021deberta} achieving strong results through token level BIO classification on standard benchmarks like CoNLL-2003~\cite{tjong2003conll} and OntoNotes 5.0~\cite{weischedel2013ontonotes}.

PII detection builds on NER but extends it to a broader set of entity types including structured identifiers such as phone numbers, email addresses, social security numbers, and credit card numbers that are not covered by traditional NER models. Microsoft Presidio
addresses this by combining rule based pattern matching for structured PII with ML based recognizers for named entities, representing a hybrid architectural approach. Presidio has seen significant adoption in industry, being integrated into Microsoft's Azure AI services and used across healthcare, finance, and government sectors for data de-identification workflows. More recently, generative LLMs have been explored for information extraction tasks, with GPT-NER~\cite{wang2023gptner} demonstrating that large language models can perform NER through prompting without task specific fine tuning. Smaller open source models such as Qwen~\cite{bai2023qwen} have made LLM based extraction accessible for production deployment where latency and cost constraints limit the use of larger models.

A key challenge across all these approaches is that PII entity types span a wide spectrum of format rigidity. Some types such as SSN and IP addresses follow strict patterns that can be reliably captured by regular expressions, while others such as person names and locations are flexible format entities whose recognition depends on contextual understanding. This spectrum has implications for OOD robustness that, to our knowledge, have not been systematically studied.

\subsection{Distribution Shift and Robustness in NLP}

The brittleness of NLP models under distribution shift has been documented across multiple tasks and domains. The WILDS benchmark~\cite{koh2021wilds} demonstrated systematic OOD performance gaps across both vision and language tasks, establishing that distribution shift is a pervasive challenge rather than a task specific anomaly. Robustness Gym~\cite{goel2021robustness} provided a unified framework for evaluating NLP model robustness along multiple axes including subpopulation analysis and transformation testing.

For NER specifically, prior work has studied robustness through controlled perturbations. Agarwal et al.~\cite{agarwal2020entity} introduced entity switched datasets to audit in domain robustness by replacing entities with alternatives from the same type, revealing that models often rely on memorized entity forms rather than contextual patterns. Lin et al.~\cite{lin2021rockner} proposed RockNER, generating adversarial examples through entity replacement from Wikidata to evaluate NER robustness. However, both approaches use artificially constructed perturbations that target individual failure modes in isolation. Real world inputs exhibit multiple simultaneous deviations (missing punctuation, informal tone, ambiguous boundaries, mixed entity contexts) that interact in complex ways not captured by single axis perturbations.

Cross domain evaluation of NER models~\cite{liu2021crossner} has shown significant performance drops when models trained on one domain (e.g., news) are applied to another (e.g., science or politics), even when the entity types remain the same. This finding is directly relevant to PII detection, where systems trained on general purpose corpora are deployed across healthcare, finance, legal, and customer support verticals with substantially different linguistic characteristics. However, cross domain NER evaluation has focused on domain vocabulary and topic shifts rather than the textual messiness and formatting irregularities that characterize real world user generated PII inputs.

\subsection{Continual Learning for NER}

A related challenge arises when NER models are iteratively improved on new data. Monaikul et al.~\cite{monaikul2021continual} studied continual learning for NER and showed that models trained on new entity types risk catastrophic forgetting of previously learned types. Xia et al.~\cite{xia2022learn} proposed reviewing synthetic samples during retraining to mitigate this forgetting effect. This line of work is directly relevant to our proposed QA driven improvement cycle (Section~\ref{sec:discussion}), where production failure cases are used to iteratively retrain models. The risk that fixing one failure category degrades performance on another makes systematic regression testing essential.

\subsection{Gaps in Existing Work}

Despite the extensive work on NER robustness, distribution shift, and continual learning, we identify a gap at their intersection specific to PII detection. No existing benchmark systematically evaluates PII detection under the natural distribution shifts characteristic of production deployment, such as unpunctuated run on text, conversational language, abbreviations, or code switching. Existing robustness evaluations use artificial perturbations that do not reflect how real users actually type. Furthermore, no prior work compares encoder based, rule based, and LLM based PII detection architectures on the same OOD benchmark to characterize their complementary failure modes. Our work addresses both gaps.

%==============================================================================
\section{Benchmark Design}
\label{sec:benchmark}
%==============================================================================

We construct a stress test benchmark comprising two evaluation sets designed to measure the gap between in distribution and out of distribution performance.

\subsection{Set A: Clean Baseline}

Set A contains 100 well formatted examples with proper punctuation, capitalization, and clear entity boundaries. We generate these using 10 manually authored templates populated with synthetic PII values from the Python \texttt{Faker} library, yielding 350 entity annotations (3.5 per example) across 9 types: \textsc{Person}, \textsc{Location}, \textsc{Address}, \textsc{Phone}, \textsc{Email}, \textsc{IP Address}, \textsc{SSN}, \textsc{Date of Birth}, and \textsc{Credit Card}. These examples mirror the formatting conventions of standard benchmarks such as CoNLL.

\subsection{Set B: OOD Stress Test}

For each of the seven distribution shift categories (Table~\ref{tab:taxonomy}), we author 8 templates capturing the characteristic shift. Each template is instantiated 10 times with different random entities, yielding 80 examples per category and 560 total with 2,330 entity annotations (4.2 per example). Gold standard span annotations are obtained automatically through the generation process since we control exactly which PII values are inserted into each template. Table~\ref{tab:benchmark_stats} summarizes the dataset statistics.

\begin{table}[t]
\caption{Distribution shift categories with representative examples.}
\label{tab:taxonomy}
\centering
\small
\begin{tabular}{lll}
\hline\noalign{\smallskip}
\textbf{Category} & \textbf{Example} & \textbf{Challenge} \\
\noalign{\smallskip}\hline\noalign{\smallskip}
Run-on text    & \textit{``my name is john smith i live at 45 oak st boston ma''} & No boundary cues \\
Conversational & \textit{``hey my buddy mike lives in nyc his email is mikej...''} & Informal register \\
Overlapping    & \textit{``I work at Amazon in Seattle and my manager is Alexa''} & Semantic ambiguity \\
Abbreviations  & \textit{``nm: John addr: 123 Main NY ph: 212-555-0123''}         & Non-standard format \\
Mixed context  & \textit{``my ip is 192.168.1.1 ssn 123-45-6789 addr 42 elm st''} & Multiple PII types \\
Typos          & \textit{``my name is jonh smth i live in san franciso''}         & Surface form noise \\
Code switched  & \textit{``my nombre is Carlos I live en New York''}              & Cross-lingual mixing \\
\noalign{\smallskip}\hline
\end{tabular}
\end{table}

\begin{table}[t]
\caption{Benchmark statistics by category. Mixed context PII has the highest entity density while overlapping entities has the lowest.}
\label{tab:benchmark_stats}
\centering
\small
\begin{tabular}{lrrr}
\hline\noalign{\smallskip}
\textbf{Category} & \textbf{Examples} & \textbf{Entities} & \textbf{Avg./Ex.} \\
\noalign{\smallskip}\hline\noalign{\smallskip}
Mixed context PII        & 80 & 450 & 5.6 \\
Run-on unstructured      & 80 & 390 & 4.9 \\
Abbreviations/shorthand  & 80 & 360 & 4.5 \\
Code switched text       & 80 & 320 & 4.0 \\
Typos/misspellings       & 80 & 290 & 3.6 \\
Conversational tone      & 80 & 270 & 3.4 \\
Overlapping entities     & 80 & 250 & 3.1 \\
\noalign{\smallskip}\hline\noalign{\smallskip}
\textbf{Total (Set B)}   & 560 & 2,330 & 4.2 \\
\noalign{\smallskip}\hline
\end{tabular}
\end{table}

%==============================================================================
\section{Experiments}
\label{sec:experiments}
%==============================================================================

\subsection{Models}

We evaluate three architectural families representing distinct approaches to PII detection (entity coverage in Table~\ref{tab:model_entities}): SpaCy \texttt{en\_core\_web\_lg}\footnote{spaCy: Industrial-strength Natural Language Processing in Python. Available at: \url{https://spacy.io/}. Accessed: July 2026.}, an encoder based general purpose NER model supporting multiple entity types (e.g., PERSON, ORG, GPE); Microsoft Presidio, a hybrid framework combining rule based (e.g., regex) recognizers for structured PII with NLP models for named entities; and Qwen2.5-3B-Instruct\footnote{Implementation available on the Hugging Face Model Hub: \url{https://huggingface.co/Qwen/Qwen2.5-3B-Instruct}. Accessed: July 2026.}, an instruction tuned LLM used with zero shot prompting for flexible extraction across entity types.

These three models were selected to cover a range of detection mechanisms rather than to represent peak performance within each paradigm. SpaCy represents the encoder paradigm where each token is classified independently based on bidirectional context learned from labeled training data. Presidio represents the hybrid paradigm where structured PII types are detected through deterministic pattern matching while named entities are handled by an underlying NER model (SpaCy in Presidio's default configuration). Qwen represents the decoder paradigm where entity extraction is framed as a text generation task, with the model producing structured JSON output in response to a natural language prompt describing the desired entity types. We deliberately chose widely available, open source tools over research prototypes because our goal is to characterize how practical, deployable systems behave under distribution shift. SpaCy and Presidio are well established open source frameworks with active communities and extensive documentation, making them common choices for organizations building PII detection pipelines. Qwen represents the class of small, locally deployable LLMs that are increasingly used where cost, latency, or data privacy constraints prevent the use of larger commercial APIs. If these readily accessible systems exhibit significant OOD degradation, the problem has immediate practical consequences for any organization relying on them.
\begin{table}[t]
\caption{Entity type coverage and detection mechanism. L = learned representations, R = regex/pattern matching, P = prompt based, (\checkmark) = partial support.}
\label{tab:model_entities}
\centering
\small
\begin{tabular}{lccc}
\hline\noalign{\smallskip}
\textbf{Entity Type} & \textbf{SpaCy (L)} & \textbf{Presidio (R+L)} & \textbf{Qwen (P)} \\
\noalign{\smallskip}\hline\noalign{\smallskip}
PERSON          & \checkmark   & \checkmark & \checkmark \\
LOCATION        & \checkmark   & \checkmark & \checkmark \\
EMAIL           & (\checkmark) & \checkmark & \checkmark \\
PHONE           & $\times$     & \checkmark & \checkmark \\
CREDIT\_CARD    & $\times$     & \checkmark & \checkmark \\
SSN             & $\times$     & \checkmark & \checkmark \\
IP\_ADDRESS     & $\times$     & \checkmark & \checkmark \\
DATE\_OF\_BIRTH & (\checkmark) & \checkmark & \checkmark \\
ADDRESS         & (\checkmark) & \checkmark & \checkmark \\
\noalign{\smallskip}\hline
\end{tabular}
\end{table}

\subsection{Evaluation Protocol}

We use relaxed span matching for evaluation: a predicted entity is considered correct if its predicted span overlaps with a gold span and the entity types match. We adopt relaxed matching rather than exact span matching because entity boundaries are often ambiguous (e.g., whether ``South Brandytown, PA'' should include the state abbreviation or not) and strict boundary requirements would penalize predictions that are functionally correct for detection purposes. We note that for downstream redaction, exact span coverage matters more, and our relaxed evaluation therefore provides an optimistic upper bound on actual redaction performance. Each model is evaluated only on entity types it fully supports (Table~\ref{tab:model_entities}). We report precision, recall, and F1 at the entity level, and define the robustness gap as $\Delta = F1_{\text{Set A}} - F1_{\text{Set B}}$, where a larger gap indicates greater sensitivity to distribution shift.
\subsection{Overall Results}

Table~\ref{tab:results} presents the overall performance and per entity type recall across both evaluation sets. All models degrade on OOD data, with the LLM achieving the highest F1 on both sets (0.921 / 0.848) but also the largest robustness gap ($\Delta = 0.073$). Presidio shows the smallest gap ($\Delta = 0.057$), benefiting from its regex components that are inherently robust to surrounding textual noise. SpaCy falls in between ($\Delta = 0.064$), limited by its narrow entity type coverage but stable on the two types it does detect.

While the LLM's overall F1 is highest, overall scores obscure critical per type weaknesses that make it unsuitable as a standalone PII solution. The LLM misses 22.3\% of person names on OOD data (recall 0.777), misclassifies nearly half of all credit card numbers as SSN (recall 0.543), and produces completely unparseable output on 2.5\% of inputs. In a safety critical context where every missed entity is a privacy violation, these are not acceptable failure rates. By contrast, Presidio achieves perfect SSN and IP detection (recall 1.000 on both sets) with zero false negatives, and SpaCy detects 95.1\% of person names on OOD data. The practical question is therefore not which single model performs best overall, but how to combine their complementary strengths to minimize the total number of missed entities across all types. A related observation is how precision and recall behave differently under distribution shift: for SpaCy and Presidio, precision remains nearly unchanged (0.678 to 0.676 and 0.755 to 0.744, respectively) while recall drops more substantially, indicating that these models become conservative on OOD inputs, missing entities rather than hallucinating them. The LLM shows a similar pattern but with an even sharper recall drop (0.917 to 0.787) while maintaining high precision (0.925 to 0.920).

\begin{table}[t]
\caption{Overall performance and per entity type recall on Set A / Set B. Each model is evaluated on entity types it fully supports (Table~\ref{tab:model_entities}). \textsc{Location} collapses across all models while rigid format entities remain robust.}
\label{tab:results}
\centering
\small
\setlength{\tabcolsep}{5pt}
\begin{tabular}{l ccc ccc c}
\hline\noalign{\smallskip}
 & \multicolumn{3}{c}{\textbf{Set A (Clean)}} & \multicolumn{3}{c}{\textbf{Set B (OOD)}} & $\Delta$ \\
\noalign{\smallskip}\hline\noalign{\smallskip}
\textit{Overall} & P & R & F1 & P & R & F1 & \\
\noalign{\smallskip}\hline\noalign{\smallskip}
SpaCy-lg      & 0.678 & 0.853 & 0.755 & 0.676 & 0.706 & 0.691 & 0.064 \\
Presidio      & 0.755 & 0.893 & 0.818 & 0.744 & 0.780 & 0.762 & 0.057 \\
Qwen2.5-3B    & 0.925 & 0.917 & 0.921 & 0.920 & 0.787 & 0.848 & 0.073 \\
\noalign{\smallskip}\hline\noalign{\smallskip}
\textit{Recall Per Type} & \multicolumn{2}{c}{SpaCy} & \multicolumn{2}{c}{Presidio} & \multicolumn{2}{c}{Qwen} & \\
\noalign{\smallskip}\hline\noalign{\smallskip}
PERSON         & 0.967 & 0.951 & 0.967 & 0.951 & 0.733 & 0.777 & \\
LOCATION       & 0.725 & 0.457 & 0.725 & 0.457 & 0.938 & 0.471 & \\
EMAIL          & \multicolumn{2}{c}{--} & 1.000 & 0.952 & 1.000 & 0.938 & \\
PHONE          & \multicolumn{2}{c}{--} & 0.800 & 0.763 & 1.000 & 1.000 & \\
SSN            & \multicolumn{2}{c}{--} & 1.000 & 1.000 & 1.000 & 0.922 & \\
IP\_ADDRESS    & \multicolumn{2}{c}{--} & 1.000 & 1.000 & 1.000 & 0.863 & \\
CREDIT\_CARD   & \multicolumn{2}{c}{--} & 0.900 & 0.700 & 1.000 & 0.543 & \\
DATE\_OF\_BIRTH & \multicolumn{2}{c}{--} & 1.000 & 1.000 & 1.000 & 0.989 & \\
ADDRESS        & \multicolumn{2}{c}{--} & \multicolumn{2}{c}{--} & 1.000 & 0.981 & \\
\noalign{\smallskip}\hline
\end{tabular}
\end{table}

\subsection{Encoder vs Decoder Architectures}

The contrast between SpaCy (encoder) and Qwen (decoder) reveals fundamental differences in how these architectures handle PII detection under distribution shift.

Encoder models like SpaCy perform token level classification where each token receives a label independently based on bidirectional context. This makes them strong at recognizing entities whose surface forms were well represented in training data. PERSON recall remains high even on OOD data (0.967 to 0.951), likely because person names have distinctive capitalization patterns and common name tokens that generalize across distributions. However, LOCATION recall drops sharply (0.725 to 0.457) because location names, especially synthetic or uncommon city names like ``Salaston'' and ``Kellieborough'', do not appear in the training vocabulary and lack the surface form regularities that person names exhibit.

Decoder models like Qwen generate entity annotations as a sequence, which allows them to leverage broader world knowledge acquired during pretraining on diverse internet text. This gives them an advantage on inputs with typos and misspellings (F1 0.869 vs 0.522 for SpaCy), as the model has likely encountered misspelled words during pretraining. However, the generative approach introduces failure modes that are absent in encoder models. First, the LLM frequently confuses entity types for ambiguous inputs: credit card numbers are labeled as SSN (recall dropping from 1.000 to 0.543), suggesting that the model struggles to distinguish between long digit sequences without explicit format cues. Second, the LLM shows lower PERSON recall than SpaCy even on clean data (0.733 vs 0.967), indicating that the zero shot prompting setup does not fully leverage the model's knowledge for this entity type. Third, the LLM produced 14 unparseable JSON outputs on Set B, representing complete detection failures where no entities are extracted at all.

In summary, encoder models fail primarily by not recognizing entities (false negatives due to unseen surface forms), while decoder models fail by misclassifying entities (type confusion) or skipping them entirely (empty or malformed outputs). Neither failure mode is strictly preferable in a safety critical PII detection context where any missed entity is a potential privacy violation.

\subsection{Per Entity Type Analysis}

The per entity type results in Table~\ref{tab:results} reveal a clear relationship between entity format rigidity and OOD robustness.

Rigid format entities such as SSN (XXX-XX-XXXX) and IP addresses (dotted decimal notation) maintain perfect or near perfect recall across both clean and OOD data for Presidio (1.000 on both sets). Their detection relies on deterministic regex patterns that match the entity format regardless of surrounding textual context. Even the LLM maintains high recall on these types (SSN: 0.922, IP: 0.863), though somewhat lower than Presidio due to occasional generation errors.

Semi-rigid format entities such as EMAIL and PHONE show moderate degradation. Email recall drops from 1.000 to 0.952 for Presidio, with failures concentrated on spelled out formats (``john at gmail dot com'') that fall outside the expected @ and dot pattern. The LLM handles these slightly worse (0.938) but for different reasons, occasionally failing to extract them in informal contexts. Phone recall for Presidio drops from 0.800 to 0.763, with failures on numbers containing extensions (``+1-766-693-8011x84895'') that fall outside standard phone regex patterns. Notably, the LLM achieves perfect phone recall (1.000) on both sets, demonstrating that the generative approach handles format variation better for this entity type.

Flexible format entities suffer the most severe degradation. LOCATION recall collapses to approximately 0.46 across all three architectures on OOD data, making it the single weakest entity type in our evaluation. This is particularly notable for the LLM (0.938 to 0.471), which shows the largest absolute drop of any entity type for any model. The LLM's failure mode is distinct: rather than missing locations entirely, it absorbs city names into broader ADDRESS spans (e.g., extracting ``588 Erickson Hills Suite 055 South Brandytown, PA'' as one ADDRESS rather than separately identifying ``South Brandytown, PA'' as a LOCATION). This represents an entity granularity problem rather than a detection failure.

CREDIT\_CARD shows an unexpected degradation pattern for the LLM, with recall dropping from 1.000 to 0.543. Analysis of failure cases reveals that the model frequently labels credit card numbers as SSN when the surrounding context is ambiguous, confusing long digit sequences that lack explicit format cues. Presidio also shows credit card degradation (0.900 to 0.700) but for different reasons: non-standard spacing and missing delimiters in messy text cause digit sequences to fall outside expected credit card patterns.

\subsection{Category Analysis}

\begin{figure}[t]
\centering
\includegraphics[width=1\textwidth]{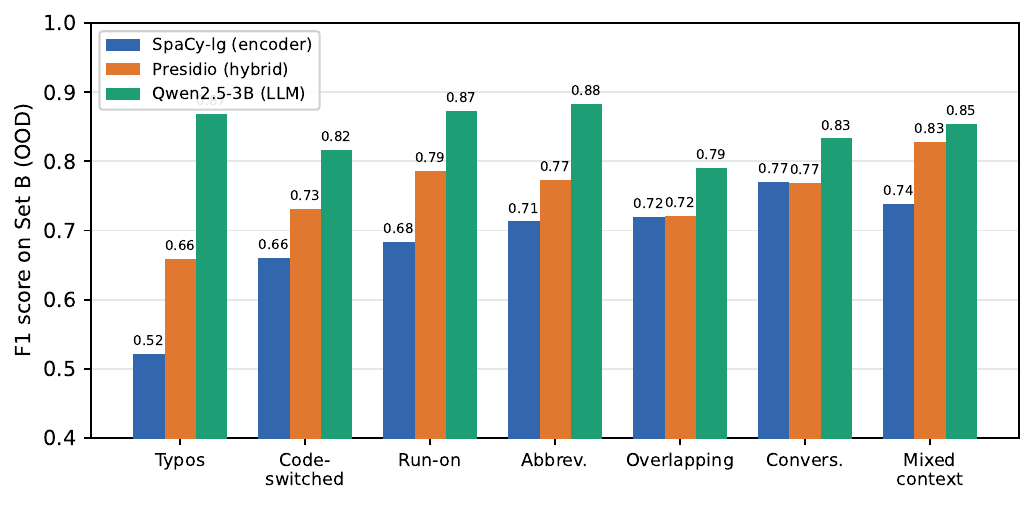}
\caption{Per category F1 on Set B (OOD). The LLM outperforms across all categories. Typos show the largest gap between architectures while overlapping entities shows the smallest.}
\label{fig:categories}
\end{figure}

Figure~\ref{fig:categories} shows per category F1 on Set B. The LLM outperforms on every category, with the largest advantage on typos (0.869 vs 0.522 for SpaCy, 0.659 for Presidio) and the smallest on overlapping entities (0.791 vs 0.721 for Presidio) where semantic ambiguity challenges all architectures regardless of their underlying mechanism.

Typos and misspellings represent the most challenging category for encoder and rule based models because both approaches rely heavily on correct surface forms. SpaCy's learned token representations break when entity text is misspelled, and Presidio's regex patterns require exact format matches. The LLM's resilience to typos (F1 0.869) likely stems from its pretraining on diverse internet text that includes misspellings, giving it a built in tolerance for surface form noise.

Run on and unstructured text (F1: SpaCy 0.683, Presidio 0.786, Qwen 0.873) poses challenges for entity boundary detection. Without punctuation and formatting cues, models struggle to determine where one entity ends and another begins. Presidio performs better than SpaCy here because its regex patterns for structured PII types can match entities regardless of sentence boundaries.

Code switched text (F1: SpaCy 0.660, Presidio 0.731, Qwen 0.817) introduces vocabulary and syntactic patterns absent from monolingual training data. Even the LLM, which was pretrained on multilingual data, shows notable degradation on this category.

Conversational and informal text yields nearly identical scores for SpaCy and Presidio (0.770 vs 0.769), suggesting that informal register alone is equally challenging for both encoder and rule based approaches. The LLM handles it better (0.833) but still shows meaningful degradation, particularly on inputs where the informal framing causes it to miss person names entirely.

\subsection{Failure Mode Analysis}

Examining individual failure cases reveals architecture specific patterns that have practical implications for system design.

\paragraph{Shared failures across all models:} \textsc{Location} entities are systematically missed when embedded in unstructured address strings. Synthetic city names generated by Faker (``Salaston,'' ``Kellieborough,'' ``Figueroahaven'') are almost never detected because they do not appear in any training data or knowledge base. Locations containing person name like tokens such as ``East David'' and ``South Melissaside'' are frequently misclassified as \textsc{Person} by SpaCy and Presidio, while the LLM absorbs them into ADDRESS spans. This shared failure on location detection, despite fundamentally different architectures, suggests a structural limitation in how current NER systems handle geographic entities in unstructured contexts.

\paragraph{Encoder specific failures (SpaCy):} Entity boundary absorption is a recurring problem in unpunctuated text. For example, ``Kimberly Snyder DVM dob'' is tagged as a single \textsc{Person} span, absorbing the professional title and the shorthand label into the name. Similarly, ``Raymond Shannon dob 09/20/1987'' merges the name with the date. Without punctuation cues, the encoder model cannot determine where one entity ends and the next begins. This boundary confusion is a direct consequence of the token level classification approach, where each token's label depends on its local context window.

\paragraph{Rule based specific failures (Presidio):} Three distinct failure patterns emerge. First, spelled out emails such as ``eugenewalker at example dot com'' are completely invisible to regex patterns that expect the @ symbol and dot notation. Second, phone numbers with extensions such as ``+1-766-693-8011x84895'' fall outside standard phone number formats. Third, the date recognizer produces excessive false positives by aggressively matching any date like numeric pattern regardless of semantic context, tagging standalone numbers like ``6574'' and ``79327'' as dates. This causes DATE\_OF\_BIRTH precision to drop to 0.492 on Set B despite maintaining perfect recall, representing a fundamentally different failure mode (over detection rather than under detection).

\paragraph{LLM specific failures (Qwen):} The LLM exhibits three distinct failure modes. First, it absorbs city names into ADDRESS spans: for input ``i live at 6766 Felicia Shore Apt. 512 Ricebury,'' it extracts the entire string as a single ADDRESS rather than separately identifying ``Ricebury'' as a LOCATION. This is architecturally interesting because the LLM recognizes the text as PII relevant but assigns it to the wrong granularity level. Second, credit card numbers are frequently misclassified as SSN when context is ambiguous (``4788861015819951'' labeled as SSN instead of CREDIT\_CARD). Third, the LLM occasionally produces completely empty outputs on conversational text, returning an empty JSON array for inputs like ``lmao yeah so Daniel Singh was like just email me at xhill@example.org'' where it apparently fails to recognize any PII in the highly informal context. It also produced 14 complete parse errors on Set B where the output was not valid JSON, representing total detection failures.

\section{Discussion and Conclusion}
\label{sec:discussion}

\subsection{Risk Assessment}

Our results show that the OOD robustness gap in PII detection is significant and pervasive across architectural families. While it may seem intuitive that models trained on clean data would struggle with messy inputs, the scale and nature of the degradation are far from obvious. \textsc{Location} recall dropping below 50\% across all architectures, including an LLM pretrained on diverse internet text, suggests a deeper structural problem than simple domain mismatch. Furthermore, the finding that different architectures fail in complementary ways on the same inputs has direct engineering implications that cannot be predicted from first principles alone.

Clean benchmark scores create a false sense of production readiness: a system reporting 0.818 F1 on clean data still misses more than half of all \textsc{Location} entities in unstructured text. Moreover, our benchmark only captures textual style shifts within a general domain. In practice, PII systems are deployed across healthcare, finance, legal, and customer support verticals where domain specific vocabulary and formatting conventions compound the distribution shift further. The robustness gaps we observe are therefore likely a lower bound on what production systems encounter.

From a risk perspective, the findings can be quantified in terms of data exposure probability. On our OOD benchmark, even the best performing model (Qwen, F1 0.848) fails to detect 21.3\% of PII entities. For an organization processing thousands of documents daily, this translates to hundreds or thousands of unredacted PII instances per day, each representing a potential regulatory violation or data breach vector. The risk is not uniformly distributed: \textsc{Location} entities have a 54\% miss rate, \textsc{Credit Card} numbers have a 30 to 46\% miss rate depending on architecture, and the LLM produces zero output (complete detection failure) on 2.5\% of inputs. Organizations that rely on benchmark F1 scores to certify their PII systems are therefore systematically underestimating their data exposure risk. This over-reliance on clean benchmark performance for deployment decisions represents a broader trust and automation risk: organizations may believe their data protection pipeline is compliant when it is in fact routinely leaking PII in real world operating conditions.

\subsection{Risk Mitigation through Hybrid Architecture}

\begin{figure}[t]
\centering
\includegraphics[width=1\textwidth]{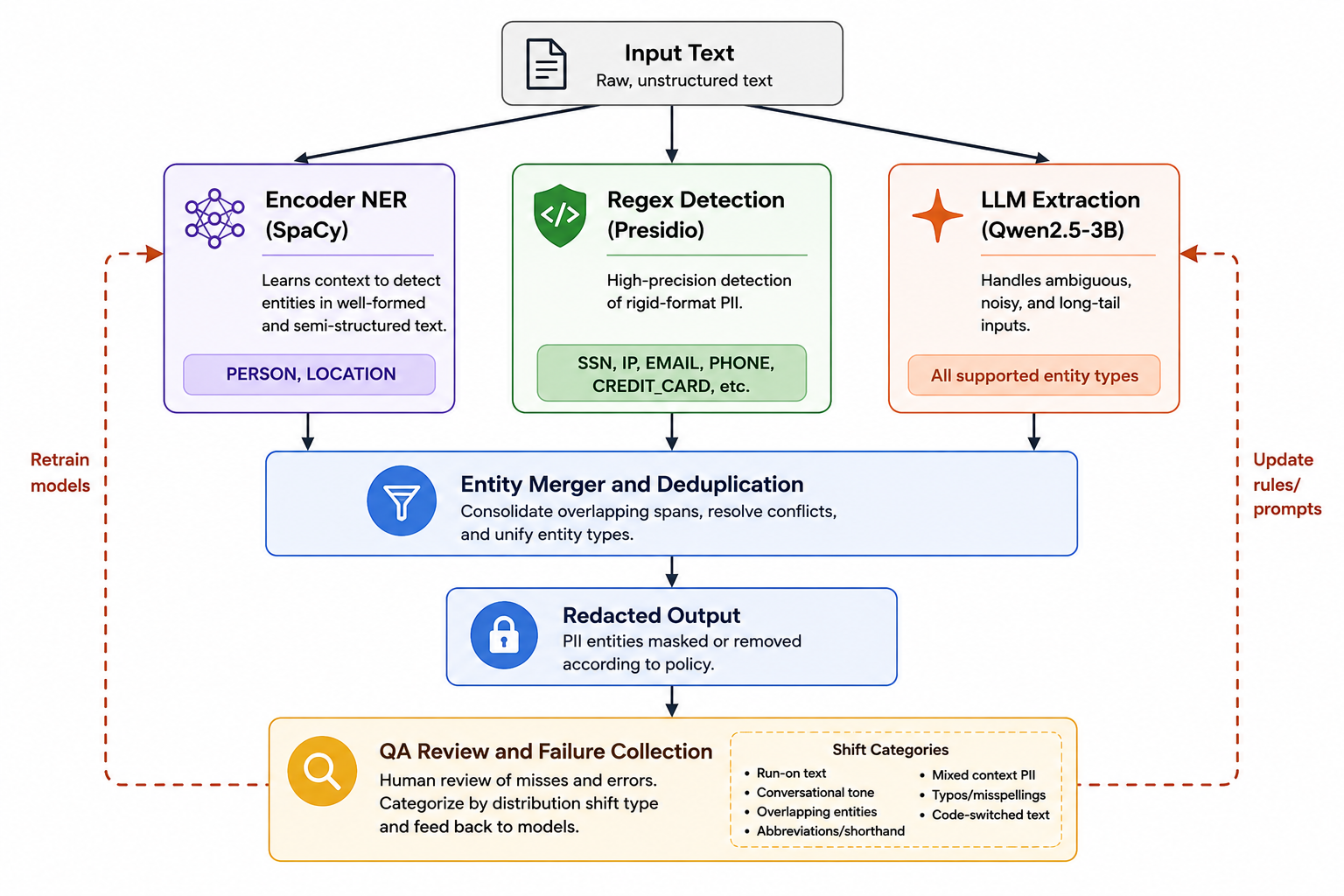}
\caption{Proposed hybrid PII detection pipeline. Three architectural families run in parallel, their outputs are merged, and a QA feedback loop iteratively improves each component based on categorized production failures.}
\label{fig:solution}
\end{figure}

Different architectures fail on different inputs, which motivates hybrid approaches (Figure~\ref{fig:solution}). Encoder models fail on unseen entity forms and boundary detection. Rule based systems fail on non-standard formats and produce false positives. LLMs achieve the best overall performance but confuse entity types and occasionally produce unparseable output. Since these failure modes are largely complementary, production systems should combine fast encoder models for well formed text, regex layers for rigid format PII, and LLM based fallback for ambiguous inputs. An entity merger layer would consolidate outputs from all three, resolving conflicts and deduplicating overlapping detections.

This hybrid approach directly mitigates the risks identified above. Regex patterns eliminate the risk of missing rigid format entities like SSN and IP addresses entirely, regardless of input messiness. The encoder model provides a fast, reliable baseline for person name detection where it outperforms the LLM. The LLM, as a generative AI component used in a defensive capacity, handles the long tail of ambiguous, informal, and noisy inputs where deterministic and encoder based approaches fail. By running all three in parallel, the system's overall miss rate becomes bounded by the intersection of their failure sets rather than the union, substantially reducing data exposure probability.

\subsection{Risk Mitigation through Iterative QA}

Beyond model architecture, we argue that a human in the loop approach grounded in software engineering practice is equally important for mitigating PII detection risk in production. Dedicated QA teams that systematically collect and categorize failure cases from production traffic play a critical role in continual risk assessment and reduction. Each failure case can be traced to a distribution shift category, and the appropriate mitigation strategy depends on the architecture. For encoder models, collected failure cases serve as additional training data for fine tuning, with targeted examples addressing specific weak categories such as typos or unseen location names. For rule based systems, QA findings translate directly into new regex patterns (for example, adding phone extension formats like ``x1234'' or spelled out email patterns). For LLMs, failures can inform prompt refinement, few shot example selection, or targeted fine tuning on the specific entity types that show the largest degradation.

These improvements are then verified through regression testing against the previously failing cases. Critically, regression testing must also cover previously passing cases, as fine tuning on new failure categories risks catastrophic forgetting~\cite{monaikul2021continual,xia2022learn} where the model improves on newly targeted inputs but degrades on inputs it previously handled correctly. Maintaining a growing regression suite organized by our taxonomy categories helps detect such regressions early and prevents the introduction of new risks while mitigating old ones. This creates a continuous risk reduction loop where each component of the hybrid pipeline improves incrementally with each deployment cycle, and the taxonomy we propose provides a structured framework for prioritizing which risks to address first based on their frequency and severity in a given domain.

\paragraph{Limitations:} Our benchmark is synthetic and operates within a single general domain. Evaluation is limited to English, and the LLM uses zero shot prompting where fine tuning could yield different results. We evaluate one representative model per architectural family; future work should include stronger encoder models (e.g., DeBERTa, GLiNER), larger LLMs (e.g., GPT-4, Llama 70B), and fine tuned variants to determine whether OOD degradation persists at higher capability levels. Future work should also incorporate naturally occurring messy text from multiple domains, expand to multilingual settings, and evaluate domain adaptive strategies for reducing the robustness gap.

\bibliographystyle{splncs04}
\bibliography{references}

\end{document}